\documentclass[
twocolumn,
]{ceurart}

\usepackage{algorithm}
\usepackage{algpseudocode}
\usepackage{tabularx}
\usepackage{hyperref}

\usepackage{amsmath,amsfonts,amssymb}
\usepackage{graphicx}
\usepackage{textcomp}
\usepackage{xcolor}
\usepackage{amsthm, algorithm, float, geometry, breakcites, enumitem, url, booktabs, arydshln}
\usepackage{mathrsfs}

\definecolor{Gray}{gray}{0.85}
\definecolor{LightCyan}{rgb}{0.88,1,1}

\newcolumntype{g}{>{\columncolor{Gray}}c}

\usepackage{array}
\newcolumntype{C}[1]{>{\centering\arraybackslash}p{#1}}

\makeatletter
\newcommand\multiline[1]{\parbox[t]{\dimexpr\linewidth-\ALG@thistlm}{#1\smallskip}}  
\makeatother
\begin{document}

\copyrightyear{2023}
\copyrightclause{Copyright for this paper by its authors. Use permitted under Creative Commons License Attribution 4.0 International (CC BY 4.0).}

\conference{Safe AI '23: Feb 13-14, 2023 | Washington, D.C., US @AAAI-23}

\title{Towards Understanding How Self-training Tolerates Data Backdoor Poisoning}

\author[1]{Soumyadeep Pal}[%
email=soumyade@ualberta.ca,
]
\cormark[1]
\address[1]{University of Alberta}

\author[2]{Ren Wang}[%
email=rwang74@iit.edu,
url=https://wangren09.github.io/,
]
\address[2]{Illinois Institute of Technology}

\author[3]{Yuguang Yao}[%
email=yaoyugua@msu.edu,
]
\address[3]{Michigan State University}

\author[4]{Sijia Liu}[%
email=liusiji5@msu.edu,
url=https://lsjxjtu.github.io/,
]
\address[4]{Michigan State University}

\cortext[1]{Corresponding author.}

\begin{abstract}
Recent studies on backdoor attacks in model training have shown that polluting a small portion of training data is sufficient to produce incorrect manipulated  predictions on poisoned test-time data while   maintaining high clean  accuracy    in downstream tasks. The stealthiness of backdoor attacks has imposed tremendous defense challenges in today's machine learning paradigm.
In this paper, we explore the potential of self-training via additional unlabeled data for mitigating backdoor attacks. We begin by making a pilot study to show that  vanilla self-training is \textit{not} effective in backdoor mitigation. Spurred by that, we propose to defend the backdoor attacks by leveraging strong but proper data augmentations in the self-training pseudo-labeling stage. We find that the new self-training regime help in defending against backdoor attacks to a great extent. Its effectiveness is demonstrated through experiments for different backdoor triggers on CIFAR-10 and a combination of CIFAR-10 with an additional unlabeled 500K TinyImages dataset. Finally, we explore the direction of combining self-supervised  representation learning  with self-training for further  improvement in backdoor defense.
\end{abstract}


\maketitle

\section{Introduction}

Deep neural networks (DNNs), key components of deep learning, have prompted a technological revolution in artificial intelligence through various applications in computer vision \cite{krizhevsky2017imagenet,goodfellow2020generative,ren2015faster} and other realms \cite{wang2020pillar,li2020short}. Due to the ever-growing capacity of DNNs, the models are capable of learning better and more accurately during the training phase. This can sometimes lead to DNNs being brittle - (1) Well-crafted imperceptible perturbations on test images can cause the model to misclassify images during the inference stage (known as evasion attack) \cite{yuan2019adversarial}. (2) Another attack called data poisoning attack \cite{goldblum2022dataset} can occur first during training by manipulating the training data by the introduction of toxic artifacts. These are memorized by the model and are carried on to the inference stage. Attack type (2) is the major focus of this work.

Recent DNNs are extremely data-hungry - they are often trained using data from anonymous or unverified sources from the internet. This makes it particularly convenient for adversaries to manipulate datasets, leading to various kinds of stealthy data poisoning attacks, thus posing a real threat to deep learning security \cite{kumar2020adversarial}. One of such data poisoning attacks is the backdoor attack (also known as Trojan attack) \cite{schwarzschild2021just,goldblum2022dataset}, where a fraction of the training data is corrupted by the addition of a trigger. In this paper, we focus on defense against such backdoor attacks. 

In many applications of deep learning, there is the availability of large quantities of unlabeled data - labeling is often cumbersome due to time and resources. Hence, semi-supervised learning \cite{4787647} has been a growing area of research, which aims to leverage such unlabeled data to improve the performance of DNNs. Self-training is one such popular paradigm, which has been proven to perform really well in large data settings \cite{xie2020self}. In this context, we aim to address the following question: 

\begin{center}
\textit{(Q) How does self-training relate to robustness against backdoor attacks ?}
\end{center}

Self-training has been recently shown to have some capability of using diverse feature priors during training \cite{pmlr-v162-jain22b}  and help alleviate spurious correlations under certain sets of assumptions \cite{chen2020self}. This inspires our study towards understanding if this paradigm may be helpful in backdoor defense. 

To the best of our knowledge, the most relevant work to ours is \cite{huang2022backdoor}. In \cite{huang2022backdoor}, self-supervised learning and symmetric cross entropy loss \cite{wang2019symmetric} is used to separate the data into probable clean and poisoned samples. Then
semi-supervised learning (MixUp \cite{zhang2018mixup}) is performed with the filtered clean samples as labeled data and the rest as unlabeled data (by removing their labels). However, this method is not able to answer our question (Q). 
Different from \cite{huang2022backdoor}, we do not aim to filter out clean samples using heuristic detection method. Though we use self-training as a semi-supervised learning algorithm for mitigating backdoor, we aim to get insights on how self-training as a sole learning paradigm can help in backdoor mitigation. We summarise our \textbf{contributions} as follows: 
\begin{itemize}
    \item We show that self-training can mitigate backdoor using additional clean unlabeled data
    \item We propose that self-training combined carefully with data augmentation has the capacity to defend against backdoor to a certain extent, even when the unlabeled data is poisoned. 
    
    \item Stronger defense is possible if stochastic data augmentation schemes (like in SimCLR \cite{chen2020simple}) are employed with self-training
\end{itemize}

\section{Related Works}

\subsection{Backdoor Attacks}
Backdoor attack is one of the emerging fields of research in data poisoning while training  neural networks. 
We focus on two types of trigger-driven backdoor attacks - poisoned label attacks and clean label backdoor attacks. 

The poisoned label attacks constitute of poisoning the training dataset by injecting a trigger in a small portion of the dataset and mislabeling them to a target class. This was fundamentally demonstrated in BadNets \cite{gu2017badnets} which used a rectangular patch and stamped it on an area of an image. Subsequently, more sophisticated triggers have been developed \cite{chen2017targeted,8802997,nguyen2021wanet,li2021invisible}. 

Another type of backdoor attack constitutes the clean label backdoor attack \cite{turner2019label,zhao2020clean}. The images belonging to the target class are adversarially perturbed away from their true class and then injected with the trigger. Training with such images establishes the correlation between the trigger and the target class. This type of attack is more stealthy because the labels of the target class are consistent with the ground truth labels. 

In this paper, we consider both types - the basic poisoned label and the clean label backdoor attack for our experiments.

\subsection{Backdoor Defense}

Due to the emerging threat of backdoor attacks, several kinds of defenses have been proposed. These roughly belong to the following categories : (1) \textit{Input preprocessing} \cite{10.1145/3427228.3427264,li2021backdoor,villarreal2020confoc,udeshi2022model}: This kind of defense introduces a preprocessing module with the intent of damaging the trigger pattern before passing it into the DNN. 
(2) \textit{Detection based defense} \cite{chen2019deepinspect,gao2019strip,tran2018spectral,kolouri2020universal,wang2019neural}: The aim here is to detect the presence of possible malicious samples or backdoored models. Then the method either denies the use of such suspicious object or filters the suspicious input samples for re-training. (3) \textit{Erasure based or model reconstruction based defense} \cite{li2021neural,zhao2020bridging,zeng2022adversarial,liu2021removing}:  This type of defense aims to erase the effect of triggers from an already backdoored model such that it performs well in both clean samples and in the presence of triggers. (4) \textit{Trigger synthesis} \cite{guo2020towards,wang2020practical,xu2020defending,chen2022quarantine}: Here the trigger is potentially detected and synthesized in the first step and then the effect of such a trigger is suppressed. (5) \textit{Poison suppression defense} \cite{du2019robust,borgnia2021strong}: This kind of defense tries to suppress the effectiveness of hidden triggers in the input samples during training, thus preventing the model from learning any correlation with the trigger. 

In this paper, we aim to suppress the poison using data augmentation and erase its effect from a trained poisoned model by self-training. Thus, our work falls within the scope of poison suppression and erasure-based defense. 

\subsection{Self-training}

Self-training is a form of semi-supervised learning \cite{4787647} which attempts to leverage unlabeled data to improve classification performance in the limited data regime. Different types of semi-supervised learning paradigms have been explored such as consistency training \cite{tarvainen2017mean,miyato2018virtual,athiwaratkun2018there,verma2019interpolation} and pseudo-labeling \cite{iscen2019label,zou2019confidence,xie2020self}. 

In self-training, a good teacher model is initially trained using the labeled data. This model is used to generate pseudolabels for the unlabeled data which are then used to train a student model. This same process is repeated iteratively. 

The main rationale behind this method is that a trained teacher model would provide better predictions on unlabeled data than pure chance. Because of the uncertainty of the correctness of predicted pseudolabels, a confidence-based example selection scheme \cite{lee2013pseudo} is often employed. Here, a fraction of pseudolabels for which the teacher model assigns the highest probability is used to train the student model. This is repeated with increasing fractions of unlabeled data till completion. 

In recent studies, self-training has been shown to have some capacity to incorporate diverse feature priors in learning \cite{pmlr-v162-jain22b}. Thus, self-training may be able to use more robust features in the data and not rely on the backdoor trigger, if designed properly. Moreover, under certain assumptions, it was shown that self-training could avoid spurious correlations \cite{chen2020self}. Thus, in this paper, we study the usefulness of self-training in mitigating stealthy backdoor attacks. 
\begin{table}[t]
  \caption{Performance of a VGG-16 model trained with self-training under different settings. The poison ratio of labeled portion of CIFAR-10 is 0.1. The model was pre-trained on poisoned labeled   portion with (SA: 81.45 \%  ASR: 100 \%)
  }
  \label{Table 1}
  \begin{tabular}{cccc}
    \toprule
    Pseudo-labeling &  $\gamma$($\mathcal{D}_U$) & SA & ASR\\
    \midrule
    \midrule
    $\mathcal{D}_U$ & Clean & 80.06 \% & 0.81 \% \\
    $\mathcal{D}_U$ &  0.1& 72.22 \% & 100 \% \\
    $\mathcal{D}_L \bigcup \mathcal{D}_U$  & Clean & 81.25 \% & 99.98 \% \\
    $\mathcal{D}_L \bigcup \mathcal{D}_U$ & 0.1 & 76.45 \% & 99.98\% \\
    \bottomrule
  \end{tabular}
\end{table}

\section{Preliminaries and  Setup}

\noindent \textbf{Backdoor Attacks.} We briefly describe the general steps to a backdoor attack. We consider a clean dataset $\mathcal{D} = \{( \mathbf{x_i}, y_i)\}_{i=1}^N$ where $\mathbf{x_i}$ is an image and $y_i$ is the corresponding label. Based on a poisoning ratio $\gamma$, the clean dataset is divided into   $\mathcal{D}_m$ and $\mathcal{D}_n$ such that  $\gamma = \frac{|\mathcal{D}_m|}{|\mathcal{D}|}$ and $\mathcal{D} = \mathcal{D}_m \bigcup \mathcal{D}_n$. $\mathcal{D}_m$ is modified with an attacker defined poisoned image generator $\mathcal{G}$ such that $\mathcal{D}_b = \{(\mathbf{x}', y_t) \ | \mathbf{x}'= \mathcal{G}(\mathbf{x}), (\mathbf{x},y) \in \mathcal{D}_m\}$. For example, one of the ways of poisoning images is to stamp a small checkerboard pattern (called trigger) at a fixed location of the image and change the labels $y$ to a target label $y_t$. Finally, the poisoned dataset $\mathcal{D}_p = \mathcal{D}_b \bigcup \mathcal{D}_n$ is sent to the users who may train a DNN on this dataset leading to the creation of a model vulnerable to backdoor attacks. 


\noindent \textbf{Threat Model.}  In this paper, we consider that the training dataset is maliciously poisoned using backdoor triggers. 
However, the user has \textit{no} prior knowledge on such train-time data poisoning. The user can obtain such a dataset, for example, by 
scraping images from the internet.
The goal of the user is to develop a training scheme to train models that are not vulnerable to backdoor attacks even at the presence of poisoned data samples.

\noindent \textbf{Problem Setup.}  Different kinds of defenses against backdoor attacks have been proposed. However, defenses based on self-training {with blind data poisoning information}    are still less explored. In this paper, we ask:
 \textit{How is self-training with additional unlabeled data useful in backdoor defense when the defender has no knowledge of backdoor attack and no access to clean samples?} 


Formally, let $\mathcal{D}_L$ be the labeled dataset and $\mathcal{D}_U$ be the unlabeled dataset which the user has at their disposal to train a model.
\textbf{We assume the worst case, where   $\mathcal{D}_L$ is always poisoned} with poison ratio $\gamma$($\mathcal{D}_L$). The unlabeled data can be clean and in the worst case, heavily poisoned. The poison ratio of the unlabeled data is denoted as $\gamma$($\mathcal{D}_U$). However, the user has no attack knowledge about any data. The model is trained using self-training with only $\mathcal{D}_L$ and $\mathcal{D}_U$. The performance of the trained model is measured in terms of \textbf{standard accuracy (SA)}, which is the benign accuracy of the model on clean samples and \textbf{attack success rate (ASR)}, which is the adversarial performance of the model on samples stamped with the train-time backdoor trigger. ASR is given by the fraction of the poisoned test samples from the non-target classes which have been predicted as the backdoor target class.

\section{Backdoor Defense Via Self-Training With Data Augmentation}

In this section, we describe our pilot study and the resultant proposed approach for defending against backdoor attacks using self-training. 

\subsection{Self-training meets Backdoor: a pilot study} 

In what follows, we present an experiment that motivates our further investigation in this direction. 

We consider a small part of the CIFAR-10 dataset \cite{krizhevsky2009learning} as labeled data $\mathcal{D}_L$ and the rest as unlabeled data $\mathcal{D}_U$.  We pretrain a model on $\mathcal{D}_L$ and perform self-training with this trained model using  $\mathcal{D}_L$ and the rest of the unlabeled data $\mathcal{D}_U$. 
Detailed experimental settings are described in \autoref{Section 5.1}. We consider the self-training algorithm described in \cite{pmlr-v162-jain22b}, which selects samples in each iteration based on their confidence levels. 

In \autoref{Table 1}, we report the performance of the model when it is self-trained with additional unlabeled data with varying poison ratio. 
The pseudo-labeling in self-training can be performed on the unlabeled data only ($\mathcal{D}_U$) or on all of the data ($\mathcal{D}_L \bigcup \mathcal{D}_U$). We \textit{eliminate} any supervisory loss from our self-training schemes because including such a loss helps in successful backdoor creation, due to the presence of backdoor triggers and malicious targets. The key insights that we get from this are as follows: 
\begin{itemize}
    \item Additional clean unlabeled data may be able to erase the backdoor effects from a poisoned model [\autoref{Table 1} row 1, ASR = 0.81\%]
    \item However, naive pseudo-labeling of poisoned data can nullify this effect. [\autoref{Table 1} row 2-4, ASR around 100\% ]
\end{itemize}
This presents the opportunity for designing a more careful self-training scheme to prevent backdoor attacks in our problem setting. 

\begin{figure*}[t]
\centering
\includegraphics[width=\textwidth]{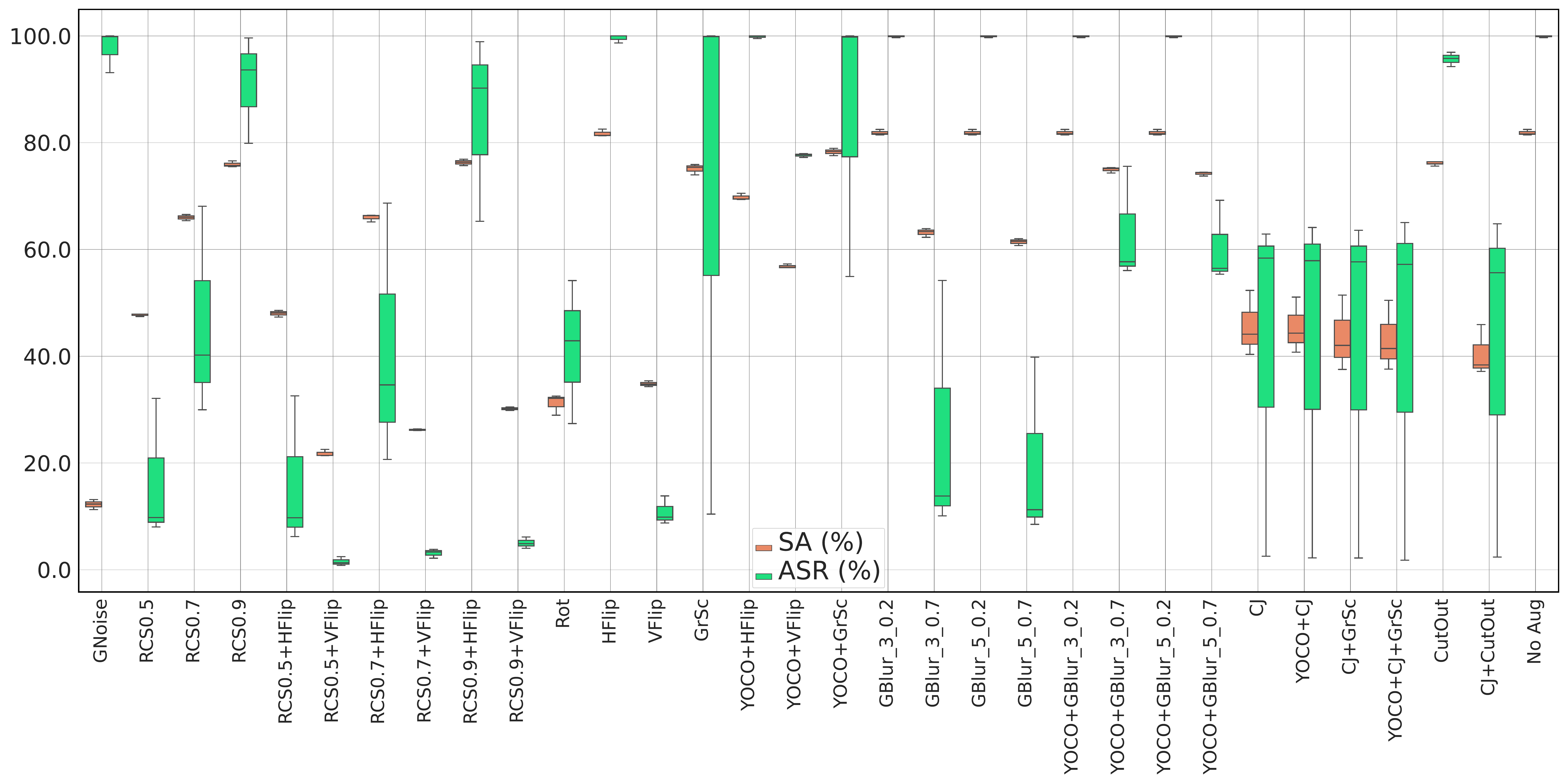}
\caption{BoxPlot for SA and ASR over different backdoor attacks with different data augmentations. Augmentations are abbreviated as follows: GNoise: Adding \underline{G}aussian \underline{Noise} with performance averaged over normal distribution of variance 0.2, 0.5, 0.7 and 1.0, RCS'x': \underline{R}andom \underline{C}ropping '\underline{x}' part of the image and re\underline{s}izing, HFlip: \underline{H}orizontal \underline{Flip}, VFlip: \underline{V}ertical \underline{Flip}, Rot: Random \underline{Rot}ation, GrSc: \underline{Gr}ay\underline{sc}ale, YOCO \cite{pmlr-v162-han22a} : \underline{Y}ou \underline{O}nly \underline{C}ut \underline{O}nce with performance averaged over horizontal cut and vertical cut, GBlur\_'x'\_'y': \underline{G}aussian \underline{Blur} with kernel size \underline{x} and standard deviation \underline{y} of the Gaussian distribution, CJ: \underline{C}olor \underline{J}itter, No Aug: No Augmentation, '+' denotes combining augmentations. }
\label{Fig. 1}
\end{figure*}

\subsection{Alleviating Backdoor Via Data Augmentation}

Spurred by the previous finding that naive pseudo-labeling in self-training cannot mitigate backdoor, we explore the landscape of data augmentations as ``feature manipulation'' mechanism to alleviate the effect of backdoor trigger. 
Data augmentations not only help in augmenting the dataset with additional data, but they also make it harder for the model to overfit to ``easy to learn but bad'' features \cite{pmlr-v162-shen22a}. This effect is especially pronounced in non-linear models like neural networks. We can think of the backdoor trigger as the ``easy to learn but bad'' feature that has a strong correlation with the target label, and hence we try to leverage the potential of data augmentations in this case. In this context, \cite{borgnia2021strong} showed that extremely strong data augmentation techniques like MixUp \cite{zhang2018mixup} can mitigate backdoor attacks in a supervised training scheme. In this paper, we explore the effect of data augmentations in self-training.

\begin{algorithm}[t]
\caption{\strut Self-training with Data Augmentation}\label{Algorithm 1}
\begin{algorithmic}
\State\multiline{\textbf{Params:}   Number of iterations N. Fraction added per iteration k.}
\State\multiline{\textbf{Input:} Labeled data $\mathcal{D}_L = \{(x_l, y_l)\}$ with $\mathcal{C}$ classes,  Unlabeled data $\mathcal{D}_U = \{(x_u, y_u)\}$, model trained on $\mathcal{D}_L$. }
\State \textbf{Data Augmentation:} $\mathcal{T}$ 
\For{iteration n $\in 1, ..., N$}
\State\multiline{forward-pass ${\mathcal{T}(x_l)}$ through model to create pseudo-labels $y_l^*$}
\State\multiline{forward-pass $x_u$ through model to create pseudo-labels $y_u^*$}
\State $\mathcal{D}_{UL} = \{(x_l, y_l^*) \bigcup (x_u, y_u^*)\}$
\State $\mathcal{D}_n$ = [];
    \For{each class c}
    \State\multiline{ Select the $\frac{kn|\mathcal{D}_{UL}|}{\mathcal{C}}$ most confident examples from $\mathcal{D}_{UL}$ predicted by the model as class c}
    \State Add those examples to $\mathcal{D}_n$ with class c;
    \EndFor
    \State Re-train (warm start) the model on $\mathcal{D}_n$ until convergence;
\EndFor
\State Train a standard model from scratch on $\mathcal{D}_N$
\end{algorithmic}
\end{algorithm}

We consider a wide variety of data augmentations to understand their effect in the presence of a backdoor trigger. VGG-16 models are trained on the labeled part of CIFAR-10 with three different types of backdoor attacks. For data augmentation, we consider rotation at different angles, adding Gaussian noise with varying variances of normal distribution, horizontal and vertical flipping, random crop and resize (with and without flip), grayscale conversion, color jitter, Gaussian blur, and CutOut \cite{devries2017improved}. We also combine suitable augmentations with YOCO \cite{pmlr-v162-han22a} to improve the diversity of augmentations. For augmentations involving stochasticity like RCS, color jitter, CutOut, Gaussiam blur and Gaussian noise, we average the results of 6 independent runs. For color jitter, we randomly choose brightness, contrast, saturation between 0.4-0.8 and hue between 0.1-0.2.

We perform the aforementioned augmentations separately on a fully poisoned test data and observe the performance of these models as shown by the box plot in \autoref{Fig. 1}. Details of the attack and training settings are included in \autoref{Section 5.1}. From \autoref{Fig. 1}, we find that there are large variances in ASR reduction across augmentations which signifies that there is no single augmentation that can combat backdoors. However, we observe that the augmentation of \underline{r}andom \underline{c}ropping of 0.5 part of the image combined with \underline{v}ertical \underline{flip}ping (RCS0.5+VFlip) reduces ASR considerably. 
We consider this particular augmentation for our future experiments.


\subsection{Self-training with data augmentations}\label{Section 4.3}

The pseudo-labeling scheme in self-training enables us to decouple any malicious targets from the training images. However, because of the strong correlation between the backdoor trigger and the given target label, pseudo-labeling would most likely predict the target label in the presence of the backdoor trigger. This could be one of the possible reasons of the high ASR in \autoref{Table 1}.

To take advantage of this decoupling phenomenon in self-training, we propose pseudo-labeling on training images with strong data augmentation. As mentioned before, we choose ``RCS0.5+VFlip'' as data augmentation. The proposed algorithm is given in \autoref{Algorithm 1}.\\ 
\noindent \textbf{Rationale.} The main rationale behind this algorithm is that pseudo-labeling a transformed backdoored image would reduce the chances of the model predicting the malicious target label, as exhibited in \autoref{Fig. 1}. However, we propose to only pseudo-label a part of the total data (in our case, we choose that to be the $\mathcal{D}_L$), to prevent a large reduction in standard accuracy. \\
\noindent \textbf{Description.} In the algorithm, we commence self-training by taking a model pre-trained on the labeled data as the teacher model. At the start of each iteration, the teacher model predicts the pseudolabels of data augmented labeled data  $\mathcal{T}(x_l)$ and unlabeled data $x_u$. For each predicted class $c$, a fraction of the most confident examples are chosen to retrain a student model. In our experiments, for each iteration, the same teacher model is used as the student model for training, and then the trained student model is treated as the teacher model in the next iteration. The fraction of the data chosen to train the model in each iteration is proportional to the iteration number. Thus, this process continues till the whole pseudolabeled data $\mathcal{D}_{UL}$ (\autoref{Algorithm 1})
is exhausted. It is important to note that we do $\textit{not}$ include any supervisory loss in our training which is usually done in standard self-training.

\subsection{Self-Training via self-supervised representation learning}

\begin{algorithm}[t]
\caption{\strut Self-training with SimCLR}\label{Algorithm 2}
\begin{algorithmic}
\State\multiline{\textbf{Params:}   Number of iterations N. Fraction added per iteration k.}
\State\multiline{\textbf{Input:} Labeled data $\mathcal{D}_L = \{(x_l, y_l)\}$ with $\mathcal{C}$ classes,  Unlabeled data $\mathcal{D}_U = \{(x_u, y_u)\}$, model trained on $\mathcal{D}_L$. }
\State\multiline{Train SimCLR representation encoder network $f(\cdot)$ with  $\mathcal{D}_L \bigcup \mathcal{D}_U$}
\State\multiline{Representation embedding $h = f(\mathcal{D}_L)$}
\State\multiline{$\mathcal{C}$ clusters $\leftarrow$ K-Means clustering of normalized $h$}
\State\multiline{$y_c \leftarrow$ Cluster Pseudolabels through Majority Voting}
\For{iteration n $\in 1, ..., N$}
\If{$n==1:$}
    \State\multiline{Predict pseudolabels $y_{ul}^*$ for $(x_l \bigcup x_u)$ using $y_c$}
\Else
    \State\multiline{forward-pass $(x_l \bigcup x_u)$ through model to create pseudo-labels $y_{ul}^*$}
\EndIf
\State $\mathcal{D}_{UL} = \{(x_l \bigcup x_u , y_{ul}^* )\}$
\State $\mathcal{D}_n$ = [];
    \For{each class c}
    \State\multiline{ Select the $\frac{kn|\mathcal{D}_{UL}|}{\mathcal{C}}$ most confident examples from $\mathcal{D}_{UL}$ predicted by the model as class c}
    \State Add those examples to $\mathcal{D}_n$ with class c;
    \EndFor
    \State Re-train (warm start) the model on $\mathcal{D}_n$ until convergence;
\EndFor
\State Train a standard model from scratch on $\mathcal{D}_N$
\end{algorithmic}
\end{algorithm}

In this section, we explore a stronger backdoor mitigation strategy using self-supervised representation learning with self-training. 

We aim to use the exemplar-based self-supervised algorithm SimCLR \cite{chen2020simple}. SimCLR is an instance discrimination based self-supervised method using contrastive loss, with instances created using stochastic data augmentation. This contrastive learning framework learns a neural network-based encoder that outputs a representation embedding of the input data.
The use of stochastic data augmentation in SimCLR creates a similar opportunity for alleviating backdoor using data augmentation and combining it with self-training. In this context, \cite{huang2022backdoor} highlighted the difference in representation space learned by SimCLR and that learned by a supervised algorithm from poisoned data, which may help in backdoor mitigation. We describe our proposed method in \autoref{Algorithm 2}.

We initially train a SimCLR representation encoder network $f(\cdot)$ as in \cite{chen2020simple} using the complete dataset. This trained representation encoder is used to find the representations embeddings $h$ of the labeled data. We cluster these embeddings into 10 (number of classes) clusters using K-Means clustering \cite{10.5555/1162264}. Since, for each of these clusters, we are aware of the ground truth labels, we pseudo-label the data in each cluster through majority voting. 

For self-training, pseudolabeling in the first iteration is done using the previously attained clusters. For any given sample $x$, we can simply find the representation vector $h_x = f(x)$ and predict its corresponding cluster by the minimum Euclidean distance between the cluster centers and $h_x$. The rest of the self-training proceeds as described in \autoref{Section 4.3}.

\section{Experiments} \label{Section5}

\subsection{Implementation details}\label{Section 5.1}

\textbf{Datasets and networks.}  We consider VGG-16 model \cite{simonyan2014very} for training using CIFAR-10.  We treat 20\% of the dataset as labeled data and consider the rest to be unlabeled. 

Additionally, we also perform a set of experiments using the complete CIFAR-10 dataset as the labeled data. For the unlabeled counterpart, we consider 80 Million Tiny Images (80M-TI) dataset \cite{torralba200880}. CIFAR-10 is a subset of this dataset - however, many images in this dataset do not belong to any of the classes of CIFAR-10. For this purpose, an unlabeled dataset of 500K images were constructed and made publicly available in \cite{carmon2019unlabeled}. Thus we use CIFAR-10 + 500K unlabeled data as our dataset and perform experiments with ResNet-18 \cite{he2016deep}. \\

\noindent\textbf{Backdoor attacks and configurations.}  We consider two main types of backdoor attacks for our experiments which are as follows: \textit{(a)} BadNet Backdoor Attack and \textit{(b)} Clean Label Backdoor Attack. For these attacks, we use a trigger of size $5 \times 5$, which is stamped at a fixed position in the images (lower right corner). The BadNet trigger can be a gray-scale patch or a RGB like \cite{saha2020hidden} trigger and the target label is always taken as 1. 

For the Clean Label Backdoor attack, images from the target class are perturbed by an adversarial perturbation so that the learned representations are distorted away from the true class. The adversarial perturbation was performed using a 10 step-PGD attack on a clean trained ResNet-18 model with the maximum perturbation $\epsilon = 8/255$ and attack learning rate $\alpha = 2/255$. The images are then stamped with a BadNet like grayscale trigger. 

The data poisoning ratio in the labeled dataset is set to be usually 10 \% for the BadNet attack and 5 \% (50 \% from the target class) for the Clean Label attack for successful poisoning. The poison ratios for the unlabeled dataset is given in \autoref{Table 2}. While using the 500K TinyImages dataset as unlabeled data,  we reduce the poisoning ratio to prevent the absolute number of poisoned samples from being too high.  \\

\noindent\textbf{Training and evaluation.}  For both pretraining and self-training, we train our models with random cropping of padding=4, random horizontal flips and random rotation of 2 degrees. We use a SGD optimizer with a momentum of 0.9 and a weight decay ratio of $1\times10^{-4}$.\\

\noindent \textit{Pretraining.} We train the models on the labeled dataset for 200 epochs with a batch size of 128. For CIFAR-10, the initial learning rate is 0.01 which is decayed by 0.5 at epoch 100. For CIFAR-10 + 500K TinyImages, the initial learning rate is 0.1 which is decayed by 0.1 at epoch 90 and 180. \\

\noindent \textit{Self-training.} We perform self-training for $N=4$ iterations and in each iteration, fraction of data added =  $0.3$. In each iteration, using the pseudolabeled dataset. In each such iteration, we train the models for 150 and 110 epochs for CIFAR-10 and CIFAR-10 + 500K TinyImages respectively. For CIFAR-10, the initial learning rate is 0.01 which is decayed by 0.5 at epoch 100, while for CIFAR-10 + 500K TinyImages, the initial learning rate is 0.1 which is decayed by 0.1 at epoch 50 and 100.

Finally, to end self-training, the model is trained from scratch using $\mathcal{D}_n$ (\autoref{Algorithm 1}, \autoref{Algorithm 2}). For CIFAR-10, this is done for 300 epochs using SGD with an initial learning rate of 0.01 which is decayed by 0.5 at 100 and 200 epochs. The corresponding training using CIFAR-10 + 500K TinyImages is performed for 250 epochs with a learning rate of 0.1 which is decayed by 0.1 at epoch 90 and 180. \\

\noindent \textit{SimCLR training.} For SimCLR, we use ResNet-18 as the base-encoder network and a 2-layer MLP projection head that produces a 128-dimensional representation space. We use the NT-Xent loss \cite{chen2020simple, oord2018representation} (with a temperature parameter of 0.5) for training SimCLR using SGD with a 0.6 learning rate, a momentum of 0.9 and  a weight decay ratio of $1\times10^{-6}$. This is trained for 1000 epochs with a batch size of 512 with standard data augmentations as used in \cite{chen2020simple}. 

\begin{table*}[htb]
  \centering
  \caption{Performance using self-training with data augmentation (\autoref{Algorithm 1}) under different settings. Semi-supervised Baseline: Self-training without data augmentation. In baseline, supervisory loss is not included for fair comparison of ASR. Poison Ratio of Clean Label Attack is the ratio of poisoned samples in the target class.
  }
  \label{Table 2}
  \begin{tabular}{C{6em}  c  c c c  c  c  c c}
    \toprule
    \multirow{2}{*}{Dataset} & \multirow{2}{*}{Backdoor Attack}&\multirow{2}{*}{$\gamma(\mathcal{D}_U)$} &\multicolumn{2}{c}{Pretrained Model}&\multicolumn{2}{c}{Semi-supervised Baseline}&\multicolumn{2}{c}{Proposed Method} \\
     \cmidrule(lr){4-5} \cmidrule(lr){6-7} \cmidrule(l){8-9}
     &  & & SA       &        ASR      &         SA       &       ASR          &  SA      &      \multicolumn{1}{c}{ASR} \\
    \midrule
    \midrule
    \multirow{5}{*}{CIFAR-10} & BadNet Gray-Scale &\multirow{2}{*}{0.1} & \textbf{81.65 \%} & 100 \% & 75.32 \% & 100 \% & 70.45 \% & \textbf{75.02 \%} \\  
    & BadNet RGB && \textbf{81.45 \%} & 100 \% & 76.45\% & 99.98 \% & 70.42 \% & \textbf{50.90 \%} \\ 
    \cmidrule(l){2-9}
    & BadNet Gray-Scale &\multirow{2}{*}{0.01} & \textbf{81.65 \%} & 100 \%  & 80.85 \% & 100 \% & 73.37 \% & \textbf{2.40 \%}\\  
    & BadNet RGB &&\textbf{81.45 \%} & 100 \%  & 80.48 \% & 100 \% & 71.49 \% & \textbf{4.98 \%} \\
    \cmidrule(l){2-9}
    & Clean-Label Attack & 0.25  & \textbf{82.45 \%} & 99.63 \%  & 81.40 \% & 98.57 \% & 73.39 \% & \textbf{44.90 \%}\\ 
    \midrule
    \multirow{3}{*}{\shortstack{CIFAR-10 + 500K \\ TinyImages}} & BadNet Gray-Scale  &\multirow{2}{*}{0.01} & \textbf{94.32 \%} & 100 \% & 89.09 \% & 99.98 \% & 84.81 \% & \textbf{14.41 \%} \\ 
    & BadNet RGB && \textbf{94.70 \%} & 100 \% & 90.19 \% & 100 \% & 84.19 \% & \textbf{14.33 \%}\\
    \cmidrule(l){2-9}
    & Clean-Label Attack & 0.05 & \textbf{93.78 \%} & 99.07 \%  & 89.43 \% & 91.19 \% & 83.67 \% & \textbf{20.61 \%}\\ 
    \bottomrule
  \end{tabular}
\end{table*}

\begin{table}[htb]
  \centering
  \caption{Performance of \autoref{Algorithm 2}. Attack used: BadNet Gray-Scale.  Dataset: CIFAR-10}
  \label{Table 3}
  \begin{tabular}{c  c c c}
    \toprule
    \multicolumn{2}{c}{Pretrained Model} &\multicolumn{2}{c}{Proposed Method}\\
     \cmidrule(lr){1-2} \cmidrule(lr){3-4} 
     SA       &        ASR      &         SA       &       ASR   \\
    \midrule
    81.65 \% & 100 \% & 71.95 \% & 0.92 \%  \\  
    \bottomrule
  \end{tabular}
\end{table}

\subsection{Experimental results}

\subsubsection{Self-training with data augmentation}

We present the performance of \autoref{Algorithm 1} in \autoref{Table 2}. The performance is measured in terms of standard accuracy (SA) and attack success rate (ASR) over different backdoor attacks. 

As mentioned in \autoref{Section 4.3}, we start self-training with a pre-trained model trained on the labeled portion of the data. We present the performance of the pre-trained model for comparison. Moreover, a semi-supervised baseline is included. The semi-supervised baseline constitutes of \autoref{Algorithm 1} without data augmentations i.e. self-training is performed only through pseudo-labeling the labeled and unlabeled data without augmentations. No supervisory loss is included in the baseline, because that would help in backdoor attack and not provide a reasonable baseline for ASR. We observe that our algorithm is successful in combating backdoor using self-training (\autoref{Table 2}: ASR of Proposed Method). 

In our experiments, we use the aforementioned strong augmentation of random cropping of 0.5 part of the image combined with vertical flipping. From \autoref{Fig. 1}, we found that the SA reduces to about $20\%$ when such augmentation is applied. However, we observe from our experiments that in our algorithm, the drop in SA is significantly less with considerable ASR reduction i.e. SA may be gained from the unlabeled data.

\subsubsection{Self-training with SimCLR}

\autoref{Table 3} presents the performance of the proposed \autoref{Algorithm 2} involving self-training with self-supervised representation learning, SimCLR. We test the effectiveness of this algorithm using BadNet GrayScale as the backdoor attack (poison ratio 0.1) on CIFAR-10. As we can see, this method is successful in improving the defense against backdoor, but it comes with a trade-off with the standard accuracy. Although only preliminary results are presented, this shows the potential of self-training integrated with SimCLR in backdoor defense.

\section{Conclusion}

In this paper, we take a step towards understanding the potential of self-training as a learning paradigm for backdoor mitigation without any available clean data and without any prior knowledge of train-time poisoning. We propose the use of strong data augmentations on part of the available data before pseudo-labeling in self-training and also explore SimCLR as a stochastic data augmentation framework in this context. We demonstrate the potential of our method across different triggers and datasets. 

Our self-training scheme, while successful in reducing backdoor, also leads to drop in standard accuracy. We attribute this to mainly the usage of strong data augmentation which leads to severe SA loss (\autoref{Fig. 1}). However, self-training is capable of recovering SA, while preserving the benefit of backdoor suppression from data augmentation. This points to the potential development of trigger-agnostic sophisticated augmentation techniques that can leverage the self-training framework to reduce ASR while maintaining SA. We hope that this work helps to motivate a deeper understanding of self-training towards its potential of backdoor mitigation, thus leading to more secure deep learning algorithms.

{\footnotesize
\bibliography{sample-ceur}}


\appendix

\end{document}